\documentclass[runningheads]{llncs}
\usepackage[T1]{fontenc}
\usepackage{graphicx}
\usepackage[dvipsnames]{xcolor}
\usepackage{amsmath}
\usepackage{todonotes}
\usepackage[caption=false]{subfig}
\usepackage{hyperref}
\usepackage{color}

\usepackage{booktabs}

\begin{document}

\newcommand{\bvalue}{\text{b}~}
\newcommand{\avalue}{\text{a}~}
\newcommand{\Lvalue}{\text{L}~}
\newcommand{\metersquare}{~$\text{m}^2$}

\title{Extended Field of View Analysis for VideoGAN-based Trajectory Generation}
\author{Annajoyce Mariani\inst{1}\orcidID{0009-0001-6647-9996} \and
Kira Maag\inst{2}\orcidID{0000-0003-1767-0476}
\and Hanno Gottschalk\inst{1}\orcidID{0000-0003-2167-2028}}
\authorrunning{A. Mariani et al.}

\institute{Technical University of Berlin, Berlin, Germany \and
Heinrich Heine University Düsseldorf, Düsseldorf, Germany}

\maketitle 
\begin{abstract}
Realistic and diverse trajectory generation is central to enabling higher levels of vehicle automation. While rule-based and classical learning-based methods may struggle to capture the complexity of traffic behavior, generative models have already demonstrated in other fields that they can handle a comparable level of complexity. 
In this paper, we build upon previous work on generative adversarial network (GAN)-based semantic bird's-eye-view traffic generation and extend the proposed framework in several key aspects. We improve the semantic representation, replace the trajectory extraction procedure with a graph-based association method, and systematically investigate increasingly larger fields of view. In addition, we introduce a quantitative evaluation framework to assess hallucinations and object permanence in generated videos. Our experiments demonstrate that the framework generalizes to larger and more complex traffic scenes while maintaining statistically realistic trajectories and coherent spatial relationships between traffic participants.
Within 150\,GPU hours of training and with inference times below 20\,ms for scenes of up to 20\,s, our results demonstrate that video-based GANs remain an efficient and scalable approach for realistic trajectory generation, even in substantially larger traffic scenes, making them well suited for downstream tasks such as prediction, planning, and simulation in automated driving.

\end{abstract}
\section{Introduction} \label{sec:intro} 
\begin{figure}[t]
    \centering
    \includegraphics[width=0.9\columnwidth]{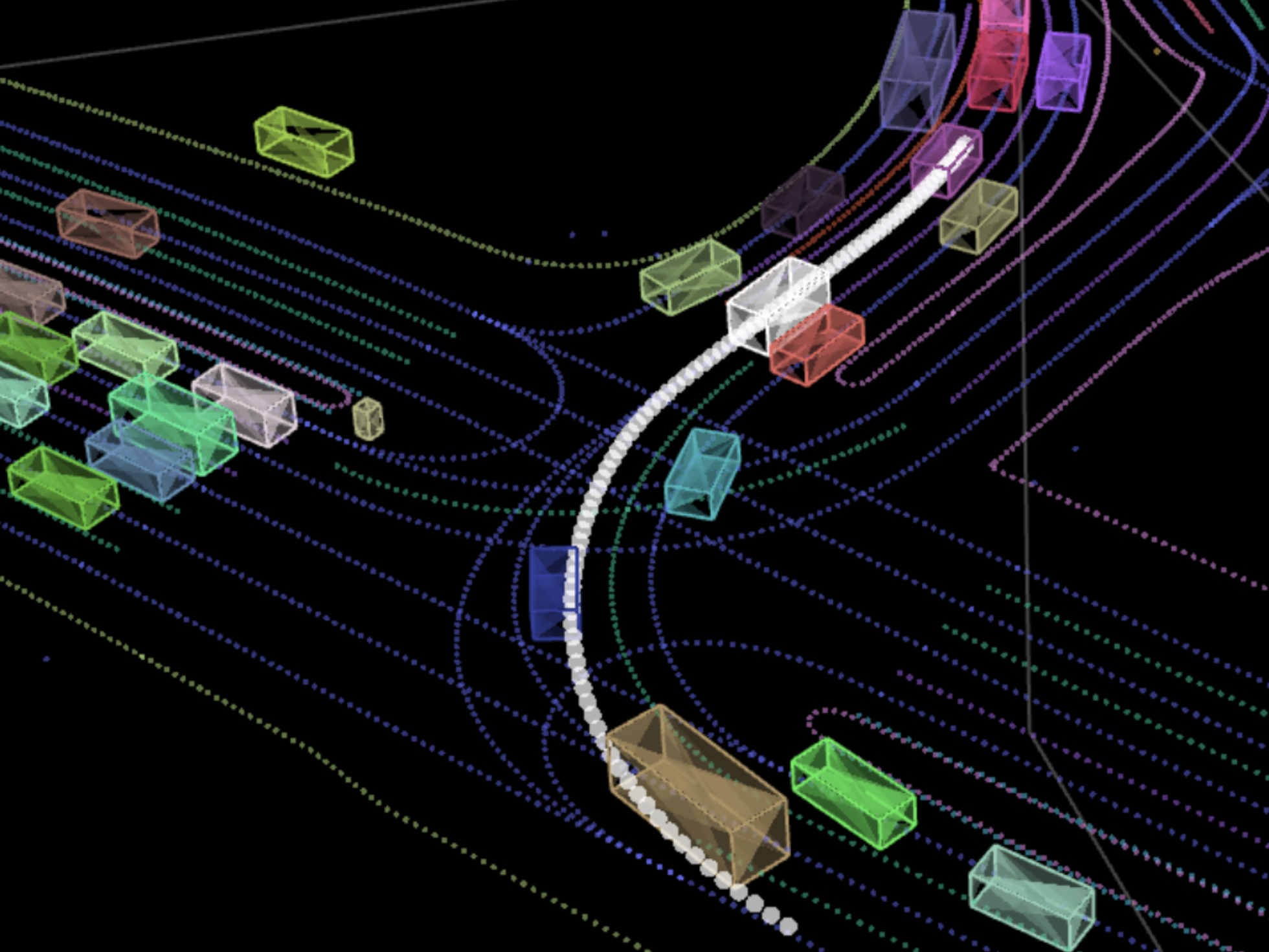}
    \caption{3D representation of a scene from the Waymo Open Motion Dataset \cite{waymo}. Image is extracted from \cite{mariani2025}.
    }
    \label{fig:example_1}
\end{figure}

Fully automated vehicles have the potential to reduce fuel consumption and traffic congestion, while increasing road safety and improving independence for humans with reduced mobility \cite{grigorescu_survey_2020}. While approaches based on explicit programming and rules could could not guarantee the amount of diversity and flexibility that characterizes the good human driver, it has become increasingly apparent in recent years that deep learning (DL) could achieve this goal \cite{huang_review_2023,khan_level-5_2023}. 
The Automated Driving (AD) task is usually decomposed into the subtasks of perception, prediction, planning, and control \cite{hu_st-p3_2022}. Among these tasks, DL has already drastically improved the capabilities of computer vision, which is employed in any AD system in the perception module \cite{horgan_vision-based_2015}. However, the real decision making center of an automated vehicle, where the versatility of DL can make a difference between stiff and hesitant driving in complex and interactive environments, lies in the prediction and planning subtasks (see Figure~\ref{fig:example_1}). 
In particular, trajectory prediction entails estimating the future positions of dynamic agents (e.g., vehicles and pedestrians), while trajectory planning aims to propose and choose trajectories compliant with the traffic regulations, the safety of other road users and the driving prompts of a human or automated navigator \cite{bharilya_machine_2023,leon_review_2021}.
The subtasks of the AD stack can either be designed and trained independently, for example, first predicting possible future paths of road users and later reason over this information to plan a safe route \cite{park_leveraging_2023,cui_lookout_2021}, or together in so called end-to-end architectures \cite{winter2025generative}.
A hybrid approach, fusing predicting and planning, consists on generating realistic futures jointly with a safe motion plan for the ego vehicle, and only intend planning as picking one of these ego trajectories compatibly with the navigation prompts \cite{fang_tpnet_2020,xiong_fine-grained_2025,zhao_novel_2022}.

In particular, two contrasting paradigms for modeling trajectories can be identified.
On the one hand, prediction is often approached as forecasting a discrete set of sequential trajectories, either with attention-based sequential networks \cite{fang_tpnet_2020,ngiam_scene_2021,zhou_hivt_2022}, or more recently with generative models, such as diffusion models or generative adversarial networks (GANs) \cite{jiang_motiondiffuser_2023,li_generative_2025}. 
These models are capable of predicting diverse and multimodal futures, as they are designed to model the full distribution of the data they are trained on by minimizing the distance of the generated distribution from the true data distribution.
On the other hand, approaches modeling scenes in a visual or semantic way, such as occupancy grids in a visual bird's-eye view (BEV) perspective, intrinsically handle spatial layout constraints such as shape-aware safety distances and joint interactions between road users and infrastructure, and contextual cues such as traffic signals and road geometry. These methods also naturally handle trajectories of variable length, such as agents entering and exiting the relevant surrounding of the automated vehicle, which can be an issue when learning trajectory sequences \cite{mahjourian_occupancy_2022,hu_fiery_2021}.
Although most of such spatially-grounded approaches make use of attention-based architectures, they do not leverage the aforementioned advantages of generative models \cite{hu_fiery_2021,gilles_home_2021,ridel_scene_2020,mahjourian_occupancy_2022}.
Prior work \cite{mariani2025} demonstrated that video generative models trained on low-resolution BEV semantical videos of traffic scenes can generate statistically accurate trajectories while capturing spatial relationships between traffic agents. 
Building upon this foundation, we extend the approach to larger and more complex scenes, while improving and widening the range of analysis and the field of view (FOV) of the generated scenes. 
Capturing a larger spatial context enables the modeling of long-range dependencies and complex multi-agent interactions, resulting in more realistic and representative traffic scenarios.
Our pipeline consists of creating semantic BEV representations of traffic scenes from abstract data, training the video-generative model and extracting the trajectories from the generated videos. We choose in particular a videoGAN model over a diffusion model due to their significantly lower computational cost, and faster training and inference times \cite{xing_survey_2025}.
While diffusion models often outperform GANs by design in terms of visual fidelity and sample diversity \cite{xing_survey_2025,saxena_generative_2021}, these advantages are less relevant for the trajectory proposal tasks, where performance is measured by task-specific metrics beyond visual quality, such as physical plausibility or driving feasibility. 
In particular, we choose a videoGAN model that has been shown to generate dynamically coherent scenes of any desired length \cite{brooks_generating_2022}.
To assess the realism of the proposed trajectories, we analyze the coherence of the generated videos, including hallucination rates, the distribution alignment of spatial and dynamic parameters such as inter-agent distances and relative speeds with respect to the input data, along with their interactions with traffic lights.
The source code of our method and the trained models are made publicly available at \url{https://github.com/ajmariani/video-gan-trajectories}.

Our contributions can be summarized as follows: 
\begin{itemize}
    \item We introduce an improved graph-based object tracking method for semantic trajectory generation, while maintaining inference times below 20\,ms for generating traffic scenes of up to 20\,s.
    \item We systematically investigate larger fields of view, evaluating scene generation over longitudinal ranges of 15\,m, 20\,m, and 25\,m.
    \item We propose a quantitative evaluation framework to assess the visual quality of generated scenes by measuring hallucinations, including disappearing and splitting vehicles.
    \item We comprehensively evaluate the realism and safety of the generated trajectories using scene density, time-to-collision distributions, relative vehicle speeds, accelerations, and interactions between traffic participants and dynamic traffic signals.
\end{itemize}

\section{Related Work}

\subsection{Deep Learning for Trajectory Proposal}

Most approaches to trajectory prediction and proposal formulate the problem as generating a discrete set of possible future trajectories for each detected agent, either conditioned or unconditioned on past observations \cite{chen_trajectory_2025}. This paradigm has been explored through encoder-decoder and self-attention architectures, including MTR~\cite{shi_motion_2022}, HiVT~\cite{zhou_hivt_2022}, Scene Transformer~\cite{ngiam_scene_2021}, and HDGT~\cite{jia_hdgt_2023}. More recently, generative models such as diffusion models and GANs have also been applied to trajectory prediction, for example MotionDiffuser~\cite{jiang_motiondiffuser_2023}, MID~\cite{gu_stochastic_2022}, and SocialGAN~\cite{gupta_social_2018}. Compared with deterministic or mode-enumerating architectures, generative models are attractive because they can explicitly model high-dimensional, multimodal future distributions and therefore produce diverse plausible futures \cite{li_generative_2025}.

An alternative line of work represents the future scene in a spatial or spatiotemporal semantic format, for example as BEV occupancy grids, occupancy-flow fields, or semantic BEV videos. Rather than enumerating a fixed set of trajectories for each agent, these methods predict scene-wise future fields. This representation naturally encodes spatial layout, static and dynamic map context, and joint interactions among road users. 
It also avoids some limitations of agent-wise trajectory representations, which typically assume a fixed set of tracked agents and fixed-length sequences, making it less natural to represent agents entering, leaving, or emerging from occlusion. This direction has been explored with recurrent models \cite{ridel_scene_2020}, convolutional networks \cite{hu_fiery_2021,wu_motionnet_2020}, and attention- or encoder-decoder-based architectures. Many of these purely occupancy-based representations struggle with instance identity and motion correspondence, especially if aiming at next-step prediction \cite{mahjourian_occupancy_2022}.
Recent occupancy-flow and implicit occupancy-field methods address or bypass this limitations by augmenting occupancy with motion information \cite{mahjourian_occupancy_2022} or limiting their aim at predicting a probability of future occupancy \cite{gilles_home_2021}.
Despite this progress, generative modeling of semantic BEV future representations remains comparatively less explored than generative modeling of agent-wise trajectory sets.
Only few recent world-model approaches have begun to generate future 3D or 4D occupancy sequences as a spatiotemporal unit with, such as OccSora, which applies transformer-based diffusion models to spatiotemporal occupancy grids \cite{11511396}.
However, diffusion models present significant inference times, which makes them more successful at scene simulation than trajectory prediction.
Our previous work \cite{mariani2025} introduced the first GAN-based frameworks for spatiotemporal trajectory generation from semantic BEV video representations. The present work builds upon and significantly extends this approach.

\subsection{Videogenerative Models}
Here we briefly compare videoGANs and video diffusion models to motivate the choice of the former.
Diffusion models generate samples by iterative denoising, often in latent space, and they are dominating the attention both in research and commercial uses \cite{singer2022makeavideo,blattmann2023stable,zheng2024opensora} due to their high visual fidelity and diversity \cite{rombach2022high}.
However, inference is notoriously expensive and wall-clock latency speeds are low and under-reported in literature, sitting in the order of magnitude of 1 to 100 seconds per scene \cite{FasterCacheLyu2025,Zhang2026_FasterDiffusion}. 
In contrast, GANs are considered less competitive than diffusion models in most video generative applications, as they generally struggle with object permanence and training instability, and they can display lower diversity than diffusion models \cite{saxena_generative_2022}.
On the other side, GAN-based models generate videos in a single forward pass, enabling significantly faster inference \cite{clark2019adversarialvideo}.
As an example, one of the benchmark models for video generation, CMD, which claims to have high inference speeds, generates a 16 frames video in 3.1 seconds \cite{EfficientCMDYu2024}, while the videoGAN model employed in this paper can generate scenes of over 1000 frames in 200\,ms, although at lower resolutions.
This efficiency makes GAN-based approaches particularly attractive for simulation and real-time applications. Advances in video generation, such as LongVideoGAN \cite{brooks_generating_2022}, have further improved the competitiveness of GAN-based video generation and demonstrated that redesigning the temporal latent representation in a StyleGAN-inspired fashion substantially improves long-term temporal consistency, object persistence, and scene diversity while maintaining efficient inference. The low-resolution branch of this architecture forms the backbone of our work, enabling the generation of dynamically coherent videos over extended horizons.
Comparing video generative models remains challenging, as no universally accepted evaluation metric exists. Common image-level metrics such as the Inception Score (IS) and Fréchet Inception Distance (FID) do not capture temporal consistency, while the Fréchet Video Distance (FVD), although specifically designed for videos, has been shown to overlook temporal artifacts and semantic inconsistencies \cite{Huang_2024_CVPR}. Consequently, these metrics only partially reflect perceptual video quality and often correlate imperfectly with human judgments \cite{unterthiner2019accurate,huang2024vbench}. This makes quantitative comparisons between different video generation paradigms, such as GANs and diffusion models, and between individual models inherently difficult.

\subsection{VideoGAN for Trajectory Generation}
In \cite{mariani2025}, a videoGAN-based trajectory proposal framework for automated vehicles was introduced, where traffic scenarios are represented as 2D semantic BEV videos. Instead of directly predicting agent trajectories, this approach learns the distribution of traffic scenes through video generation.
Abstract trajectory data are rasterized into sequences of low-D semantic frames using an HSV-based color encoding, which represent different scene elements such as road structures, traffic lights, and vehicles. A videoGAN is then trained on these semantic sequences to generate realistic future traffic scenarios. The generated trajectories are recovered by extracting object positions from the generated frames and associating object identities across timesteps using a cost matrix-based matching procedure. This formulation allows the model to capture both spatial relationships and temporal dynamics between multiple traffic participants.

In this work, we build upon this framework and extend it in three main aspects. First, we replace the HSV-based representation with a Lab color-space encoding to improve the separation between semantic categories and facilitate the extraction of objects and trajectories from the generated scenes. Second, we introduce a graph-based trajectory extraction approach, where detected objects in individual frames are represented as nodes and temporal associations are established through edges based on the reciprocal overlap of objects in adjacent frames. Finally, we extend the generation framework to larger FOVs, enabling the modeling of more extensive traffic contexts and long-range interactions between agents.

\subsection{Metrics for Trajectory Proposal}
Two main types of metric exist to assess safety and realism of generated trajectories: displacement metrics, and statistical scores. 
Displacement metrics, such as Average Displacement Error and Final Displacement Error, measure how close a predicted trajectory is to its corresponding ground truth.
These metrics are not meaningful in our framework, as our model does not predict continuations of observed trajectories but instead generates entirely new scenes of traffic by learning distribution of the training data.

Statistical scores assess the safety and realism of generated trajectories in terms of red-light violation, collision and overlap rates, unsafe spacing, and time spent driving off-road or in the wrong lane \cite{huang_multimodal_2023,rhinehart_deep_2019}.
The time-to-collision is also used to assess the level of risk of a scene even if no collision happen, by measuring the ratio between the distance between two vehicles and their closing speed, i.e., their speed relative to each other \cite{beyondLi2024}.
Driving smoothness and realism are also evaluated statistically by comparing the distributions of parameters such as vehicle speed, acceleration, jerk and inter-vehicle distance \cite{jiao_kinematics-aware_2024}.
We use a similar statistical comparison between the speeds and accelerations of the generated and training data.

To capture realism and physical plausibility in the trajectory proposal context, we therefore adopt statistical distributional comparisons between generated and real data in quantities such as speeds and accelerations relative to the ego vehicle, the average distance and time-to-collision between vehicles, and their speeds around red and green traffic lights, along with static analysis of scene density and object sizes.
We complement the statistical evaluation with a task-specific visual metric that quantifies hallucinations in generated traffic scenes by measuring the occurrence of appearing, disappearing, merging, and splitting vehicles.

\begin{figure*}[ht]
   \centering
   \includegraphics[width=\textwidth,height=20cm,keepaspectratio]{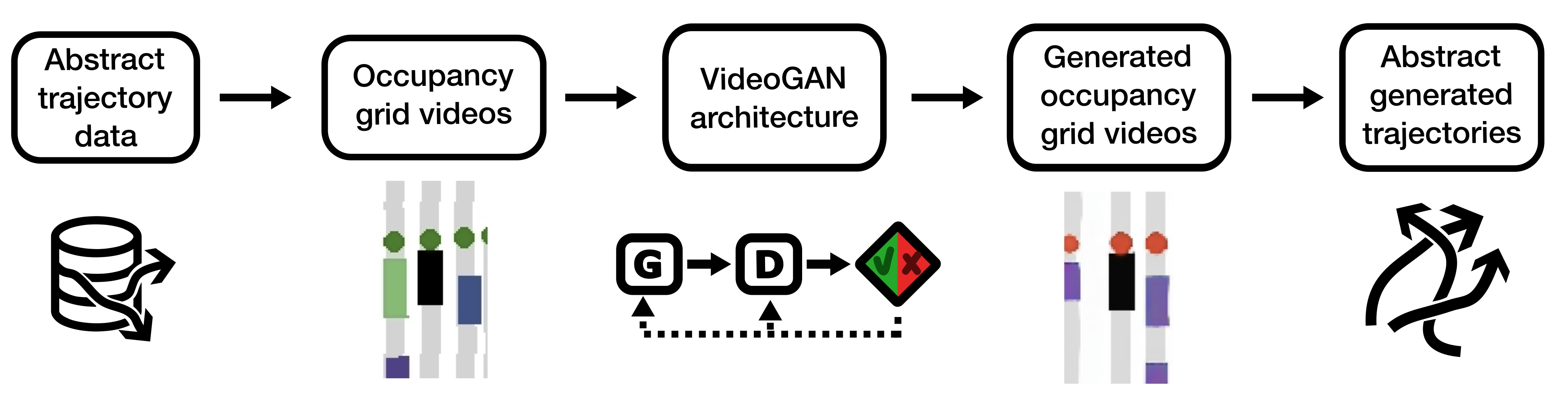}
   \caption{A schematic illustration of the pipeline. The abstract trajectory data is rasterized into low-resolution BEV semantic videos of traffic scenes. A videoGAN model, which consists of a generator (G) that attempts to generate realistic data, and a discriminator (D) that distinguishes real data from generated data, is trained on the semantic scenes. Last, abstract trajectories are extracted from the generated videos. Illustration adapted from \cite{mariani2025}.}
   \label{fig:method}
\end{figure*}

\section{Proposed Method}
In this section we present the data processing and training pipeline we use to create semantic BEV representation of the traffic scenes from an AD dataset, train a videoGAN model on them, and extract trajectory data from the generated videos. A schematic representation is shown in Figure~\ref{fig:method}. 

\subsection{Rasterization of Trajectory Data}
The trajectory data from a street scene dataset is first converted into a semantic raster representation. 
This representation captures the spatial layout of the traffic scene frame by frame, while preserving the temporal evolution of the traffic participants across the full scene. 
Our semantic representation of traffic scenes includes three elements for each timestep:
\begin{enumerate}
    \item centerlines of the road map,
    \item position, orientation and two-dimensional shape of the ego car and the agents in its surrounding, i.e., in its FOV,
    \item state and position of traffic lights in the FOV. 
\end{enumerate}

First, we convert all the world coordinates into image coordinates, where each scene is centered around a designated reference agent from a BEV perspective.
For real-world datasets, this reference is naturally chosen as the recording vehicle, as observations typically become less reliable with increasing distance from the sensor platform. In contrast, synthetic datasets allow any traffic participant to be selected as the scene center, substantially increasing the number of training samples that can be extracted from a single simulation, depending on the chosen FOV.
\begin{figure}[tb]
    \centering
    \begin{minipage}[t]{0.45\textwidth}
        \centering
        \includegraphics[width=\linewidth]{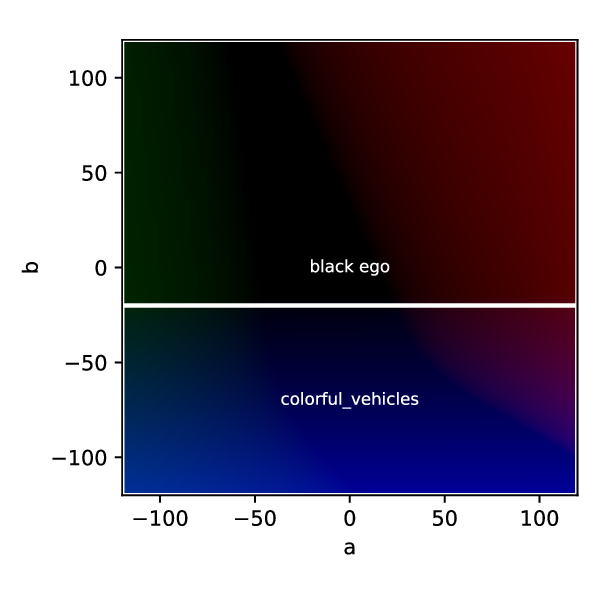}
        \vspace{3ex}
        {\small \Lvalue=-10}
    \end{minipage}
    \hspace{0.4em}
    \begin{minipage}[t]{0.45\textwidth}
        \centering
        \includegraphics[width=\linewidth]{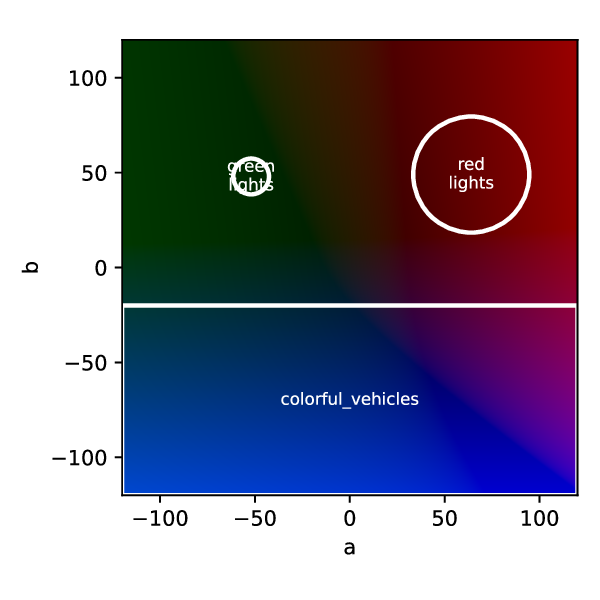}
        \vspace{1ex}
        {\small \Lvalue=10}
    \end{minipage}
    \hspace{0.4em}
    \begin{minipage}[t]{0.45\textwidth}
        \centering
        \includegraphics[width=\linewidth]{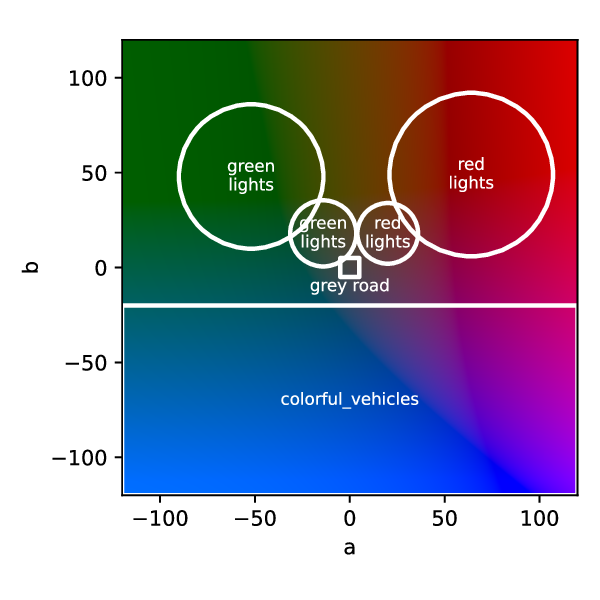}
        {\small \Lvalue=30}
    \end{minipage}
    \hspace{0.4em}
    \begin{minipage}[t]{0.45\textwidth}
        \centering
        \includegraphics[width=\linewidth]{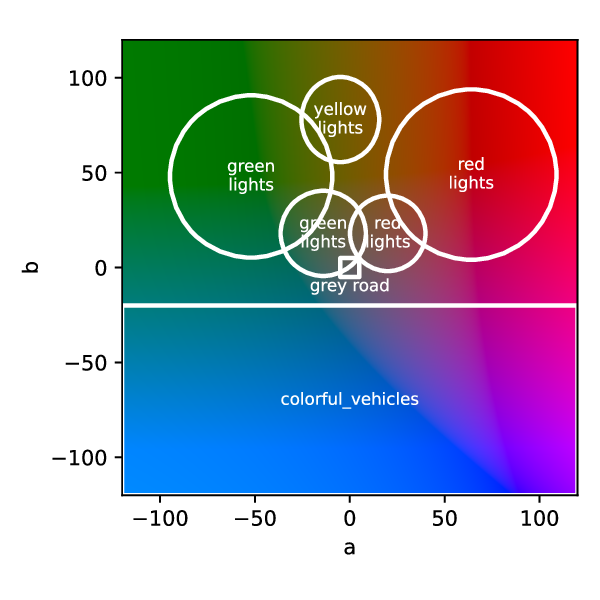}
        \vspace{1ex}
        {\small \Lvalue=40}
    \end{minipage}
    \caption{Four slices of the Lab space represented in RGB.}
    \label{fig:Lab_partitions}
\end{figure}
Each element of the traffic scene is assigned a three-channel color value. Although the videoGAN could, in principle, learn arbitrary value combinations and channel representations independent of their RGB interpretation, GANs are known to exhibit unstable training behavior in practice. Since we do not identify a compelling advantage of using a non-color encoding, and the employed videoGAN has demonstrated strong performance on RGB video generation \cite{brooks_generating_2022}, we adopt an RGB-based representation throughout this work.
To preserve the semantic identity of the different scene elements, we employ a color encoding that maximizes the separation between semantic categories. This facilitates both the learning process of the videoGAN and the subsequent extraction of semantic information from the generated videos.
Moreover, assigning visually distinguishable colors to semantic objects provides a simple qualitative means of visual assessing training progress.

\begin{figure}[t]
    \centering
    \subfloat[15\,m real]{
    \includegraphics[width=0.15\columnwidth]{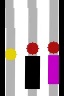}
    }
    \hspace{0.3ex}
    \subfloat[20\,m real]{\includegraphics[width=0.15\columnwidth]{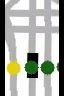}}
    \hspace{0.3ex}
    \subfloat[25\,m real]{  \includegraphics[width=0.15\columnwidth]{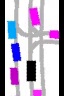}}
    \hspace{1.3ex}
    \subfloat[15\,m gen.]{\includegraphics[width=0.15\columnwidth]{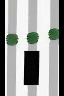}}
    \hspace{0.3ex}
    \subfloat[20\,m gen.]{\includegraphics[width=0.15\columnwidth]{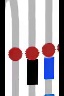}}
    \hspace{0.3ex}
    \subfloat[25\,m gen.]{\includegraphics[width=0.15\columnwidth]{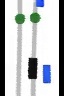}}
    \caption{Some examples of frames from real traffic scenes (a,b,c) and from generated ones (d,e,f) for the three FOVs investigated in this work. The scenes are represented as occupancy grid videos in BEV perspective, with road users as 2D Bounding Boxes, traffic lights as green, yellow or red circles, and road as gray center lines.}
    \label{fig:example_2}
\end{figure}

The semantic representation comprises the following categories: empty space, road markings, the three traffic light states, the ego vehicle, and other vehicles.
We found the Lab color space to be particularly well suited for clearly separating these semantic categories. 
Neither the RGB nor the HSV color spaces (used in \cite{mariani2025}) provide the same degree of separation, whereas Lab enables a more distinct categorization while remaining easily interpretable by humans. The Lab color space represents colors using three channels: \Lvalue for lightness, \avalue for the green-red axis, and \bvalue for the blue-yellow axis. Some 2D slices of the 3D Lab space, rendered in RGB, are displayed in Figure~\ref{fig:Lab_partitions}.
Figure~\ref{fig:example_2} shows some examples of semantic traffic frames at different FOVs, both rasterized from real data (on the left) and generated by the model (on the right).
Center lines are rendered in the background as continuous lines of specified thickness and gray color, corresponding to the Lab range where \Lvalue$\in$~[30,90], and \avalue and \bvalue values are around 0. In our experiments, a thickness of 1.5\,m is used for the center line.
Traffic lights are rendered as red, green, or yellow circles of size 2$\times$2\,m in the foreground.
Two ellipsoidal ranges are selected to cover red and green respectively, and one to cover yellow, as displayed in Figure~\ref{fig:Lab_partitions}.

Each road user is rendered as a rectangle of the specified size, position and orientation. The ego car is black (low \Lvalue levels), while other road users have colors sampled from the remaining blue-pink range, characterized by a negative \bvalue value, and positive \Lvalue.
After rasterization, each frame is cropped to window of specified pixel size and aspect ratio that represents the surroundings of the ego vehicle.

\subsection{Training}
We follow the training procedure proposed in the previous work \cite{mariani2025}, as it has been shown to produce high-quality traffic scene generations using a video GAN. By adopting the same training strategy, we ensure a fair comparison while allowing our extensions to be evaluated independently of changes to the underlying generative model.

\subsection{Frame by Frame Object Detection}
The first step in the extraction of trajectories from a generated (or training) video is identifying the objects present in each frame.
In order to do so, a pipeline mostly based on color filtering in the Lab space is used. The Lab ranges associated to each semantic categories are the same used to rasterize the training data, as described above.
First, a filter is applied which produces a mask from red, green and yellow pixels respectively.
Then, these pixels are inpainted in the original frame, such that a vehicle passing below a traffic light can be better detected as a single object.
A similar masking is applied to the black ego vehicle and to the other road users.

\begin{figure}[tb]
    \centering
    \includegraphics[width=0.11\textwidth]{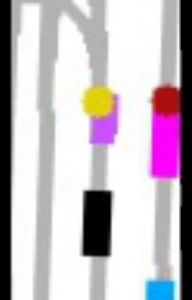}
    \includegraphics[width=0.11\textwidth]{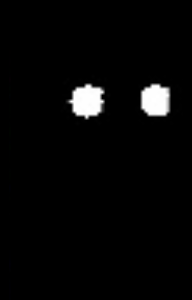}
    \includegraphics[width=0.11\textwidth]{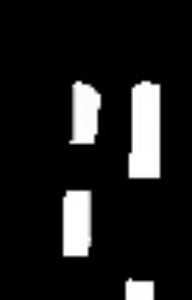}
    \includegraphics[width=0.11\textwidth]{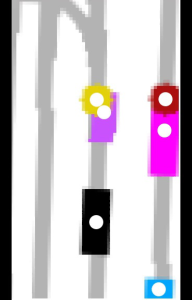}
    \hspace{3ex}
    \includegraphics[width=0.11\textwidth]{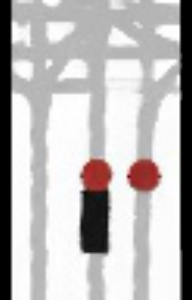}
    \includegraphics[width=0.11\textwidth]{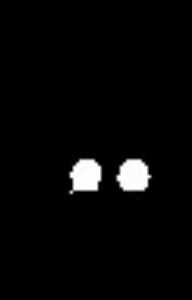}
    \includegraphics[width=0.11\textwidth]{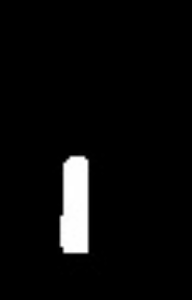}
    \includegraphics[width=0.11\textwidth]{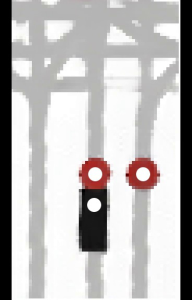}
    \caption{Two sets of frames displaying the frame by frame extraction process and accuracy. The first set is from a real video, the second from a generated video. For each set, moving left to right, we see: the original frame, the mask for traffic lights, the mask for vehicles, and original image overlaid with the pipeline prediction of the center of the object. The object detection pipeline is fairly accurate even when a vehicle is similar in color to or passing beneath a traffic light. In the pipeline, the masks for the ego agent and other road users are separated, but they are merged in this figure for visualization purposes. Notice that due to the inpainting step, a vehicle adjacent to a traffic light can appear as slightly longer in the extracted mask.}
    \label{fig:frame_by_frame_decoding}
\end{figure}

Each of the masks is analyzed to extract spatial features, including center coordinates, size, color, area, and shape descriptors such as rectangularity, circularity and aspect ratio.
Figure~\ref{fig:frame_by_frame_decoding} displays two sets of frames and mask, one on a real frame, one on a generated one. For each set the original frame, the traffic light mask, the vehicles mask, and the original frame overlapped with the identified center of each of these objects are shown.
Both the extracted mask and the deduced centers show that this step of the pipeline is very effective at identifying objects from the semantic representation, with high accuracy.

The output of this pipeline is a list of objects with the aforementioned attributes, along with the corresponding timestep and the mask category from which they were derived.

\subsection{Trajectory Extraction}\label{subsec:trajectory_extraction}
\begin{figure}[tb]
    \centering
    \includegraphics[width=\columnwidth]{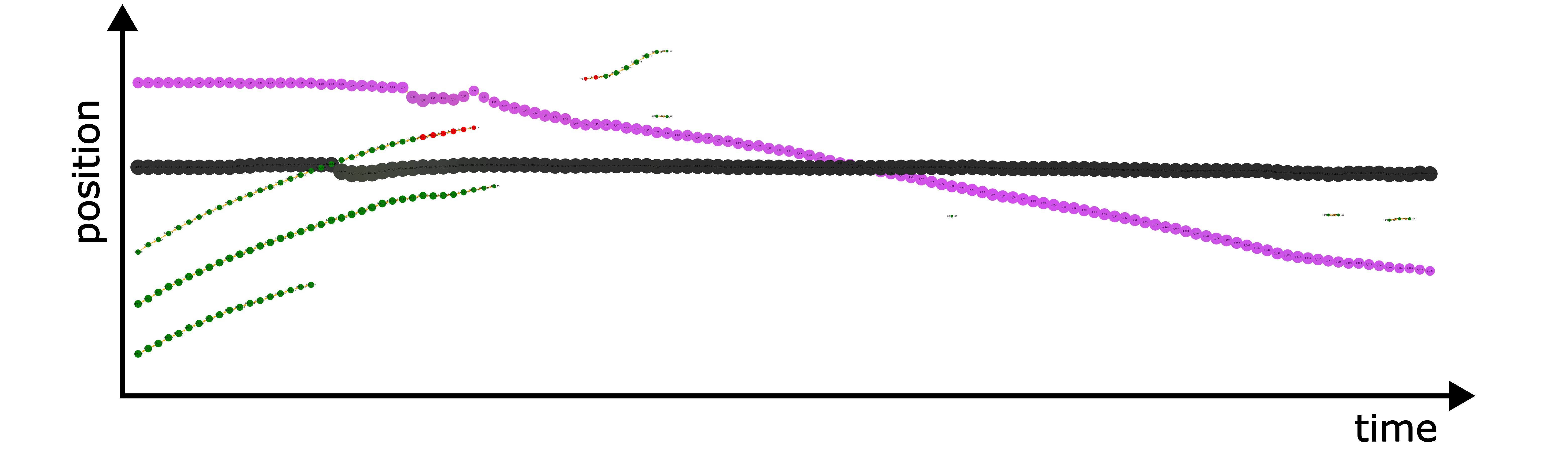}
    \caption{An example of trajectory graph extracted from a real video. The y-axis represents a combination of longitudinal and lateral position, such that higher distances between two objects are represented by higher distances in position. The black line in the center corresponds to the ego vehicle, whose position remains stable in the center of the scene across time. In the beginning of the scene, three green light are present, and both the black and the purple vehicle move across them. Later in the scene, the purple agent is overtaken by the the ego vehicle. Notice one of the traffic lights turning right after the ego vehicle has crossed it.}
    \label{fig:graph_trajectory_generated}
\end{figure}

Managing to extract trajectories from the generated videos is a crucial point in a video-generative approach to trajectory proposal.
In this work, we follow a graph-based approach, where each object in each frame is assigned to one node, with properties such as their position, their object type, and their timestep. Then, we construct a graph by connecting each node to each other node in the following frame which has the same object type, and whose spatial positions overlap in any measure. The intensity of the edge is proportional to the overlap.
Because the time difference between two frames is only 100\,ms, even in the larger investigated FOV of 25\,m, the amount of pixels that a vehicle can cover in the 100\,ms elapsing between one frame and the next is lower than the size of the car. Namely, a 96\,pixels frame representing of 25\,m FOV corresponds to around 3.8\,pixels per meter. If a vehicle was to move at the excessive speed of 90\,km/h relative to the ego vehicle, their position would still only change of 2.5\,m in 1000 ms, which is less than the average vehicle length of 4.5\,m.
This is not always true for traffic lights, for which an additional mechanism is applied to link isolated nodes, if they are close enough in position, although not overlapping.
A representation of the trajectory graph is shown in Figure~\ref{fig:graph_trajectory_generated}. 
From the trajectories so extracted, position, speed and acceleration of each object relative to the ego vehicle can be computed.

\section{Experiments}
\subsection{Experimental Setting}

\begin{figure}[tb]
    \centering
    \begin{minipage}[t]{0.11\textwidth}
        \centering
        \includegraphics[width=\linewidth]{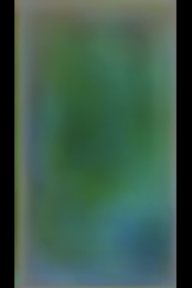}
        \vspace{3ex}
        {\small 0 ep.}
    \end{minipage}
    \hspace{0.1pt}
    \begin{minipage}[t]{0.11\textwidth}
        \centering
        \includegraphics[width=\linewidth]{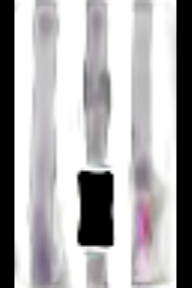}
        \vspace{3ex}
        {\small 1.5 ep.}
    \end{minipage}
    \hspace{0.1ex}
    \begin{minipage}[t]{0.11\textwidth}
        \centering
        \includegraphics[width=\linewidth]{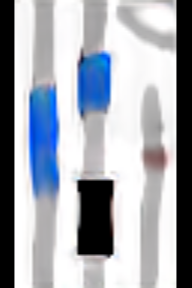}
        \vspace{1ex}
        {\small 2.5 ep.}
    \end{minipage}
    \hspace{0.1pt}
    \begin{minipage}[t]{0.11\textwidth}
        \centering
        \includegraphics[width=\linewidth]{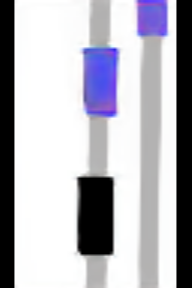}
        {\small 4 ep.}
    \end{minipage}
    \hspace{0.1pt}
    \begin{minipage}[t]{0.11\textwidth}
        \centering
        \includegraphics[width=\linewidth]{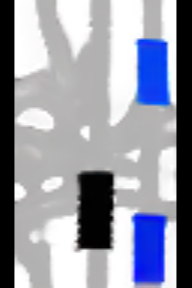}
        \vspace{1ex}
        {\small 9 ep.}
    \end{minipage}
    \hspace{0.1pt}
    \begin{minipage}[t]{0.11\textwidth}
        \centering
        \includegraphics[width=\linewidth]{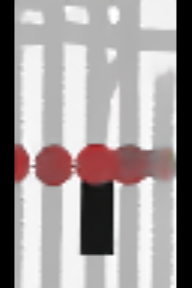}
        {\small 17 ep.}
    \end{minipage}
    \hspace{0.1pt}
    \begin{minipage}[t]{0.11\textwidth}
        \centering
        \includegraphics[width=\linewidth]{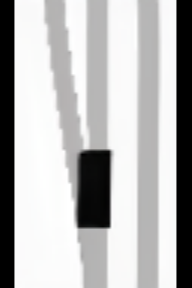}
        {\small 20 ep.}
    \end{minipage}
        \hspace{0.1pt}
    \begin{minipage}[t]{0.11\textwidth}
        \centering
        \includegraphics[width=\linewidth]{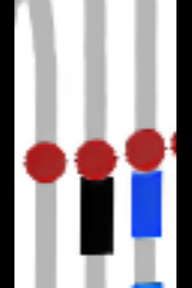}
        \vspace{1ex}
        {\small 23 ep.}
    \end{minipage}
    \caption{Examples from progressing training iterations at FOV 20\,m}
    \label{fig:train}
\end{figure}

We train the low-resolution network from LongVideoGAN \cite{brooks_generating_2022} on a 80GB A100 GPUs for 140\,GPU hours on 60k video sequences of 128 frames (12.8 seconds) each, for a total of around 260\,h of driving content. We train for 200k steps with a batch size of 8, corresponding to around 26 epochs.
We visually evaluate the progress of the training, and we test quantitatively over 200 scenes of 128 frames.
We investigate rectangular scenes of pixel size 54$\times$ 96, with three different FOVs: 15\,m, 20\,m and 25\,m in the longitudinal direction.
The training videos are produced from the Waymo Open Motion Dataset~\cite{sun_scalability_2020}, one of the largest and most diverse driving datasets.
Inference takes approximately 20\,ms for a 15\,s scene and scales linearly with sequence length, reaching 150\,ms for a 2-minute scene. 
Example images of training progress over iterations are shown in Figure~\ref{fig:train}.

\subsection{Quality of the Generated Videos}
\begin{figure}[tb]
    \centering
    \subfloat[15\,m]{\includegraphics[width=0.15\textwidth]{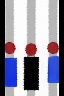}}\hspace{0.6em}
    \subfloat[20\,m]{\includegraphics[width=0.15\textwidth]{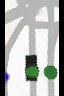}}\hspace{0.6em}
    \subfloat[20\,m]{\includegraphics[width=0.15\textwidth]{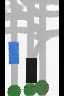}}\hspace{0.6em}
    \subfloat[15\,m]{\includegraphics[width=0.15\textwidth]{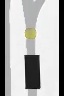}}\hspace{0.6em} 
    \subfloat[25\,m]{\includegraphics[width=0.15\textwidth]{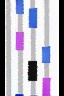}}\hspace{0.6em}
    \caption{Significant frames from generated videos of traffic scenes of different fields of view. The agents are observed waiting at a red light (a), crossing (b) and passing (c) a green light, approaching a yellow light (d) and moving in dense traffic (e)}
    \label{fig:frame_examples}
\end{figure}

At \url{https://www.youtube.com/watch?v=7BAulb8rGsU} we provide a selection of real and generated video clips for the three different FOV sizes of 15\,m, 20\,m and 25\,m. Moreover, Figure~\ref{fig:frame_examples} shows some examples in different FOVs of typically observed situations, such as passing a green light, waiting at a red light or navigating dense traffic.
We find the best checkpoint at 20 epochs for the 15\,m run, at 23 epochs for the 20\,m run, and at 24 epochs for the 25\,m run.
In all three cases, we can see that the generated scenes look qualitatively very similar to the training data in colors and features.
The generated vehicles drive well-aligned with the center lines, without sticking rigidly to the center, and their dynamics are smooth, without jerks in relative acceleration.
We do not observe issues with object permanence, which can sometimes affect video-generative models. On the contrary, we observe instances of an agent leaving the frame, and later reappearing in a dynamically consistent position, which suggests the model is able to model these situations. 
Moreover, we do not observe any statistically relevant instance of agents spawning, merging or splitting, although occasionally, especially in larger scenes of FOV 25\,m, some vehicles are observed to slowly vanish at red traffic lights for long scenes.
The only visual inconsistencies affect the behavior of traffic lights, which are sometimes observed to morph slightly when a car is passing under them, and the occasional color shift of a non-ego car along the scene. Neither of these inconsistencies affect trajectory reconstruction.

In addition to this, the generated videos feature diverse and accurate behavior of the ego agent with respect to the traffic lights.
Lights of the three colors are observed, and the generated samples also display dynamic color changes. We find that most of the times the agents proceed when the light is green, but slows down and stops at the red light.
Only one situation that is observed sometimes in real videos was never observed in the generated ones, namely the ego car driving on a narrow road flanked by parked agents. We believe this is not learned by the videoGAN due to being a rare occurrence and visually very different from usual scenes. However, parked cars do not seem to be very relevant to the behavior of vehicles driving on the road. The road lines also appear mostly realistic in size and behavior, although intersections are often represented as a confused web of center lanes, especially in scenes of lower FOV.
In all cases, the agents appear to keep a safety distance. All these observations are valid throughout videos of any length, up to 2 minutes long, for the three FOVs investigated.

\subsection{Quantitative measure of the hallucinations}
\begin{figure}[tb]
    \centering
    \includegraphics[width=\columnwidth]{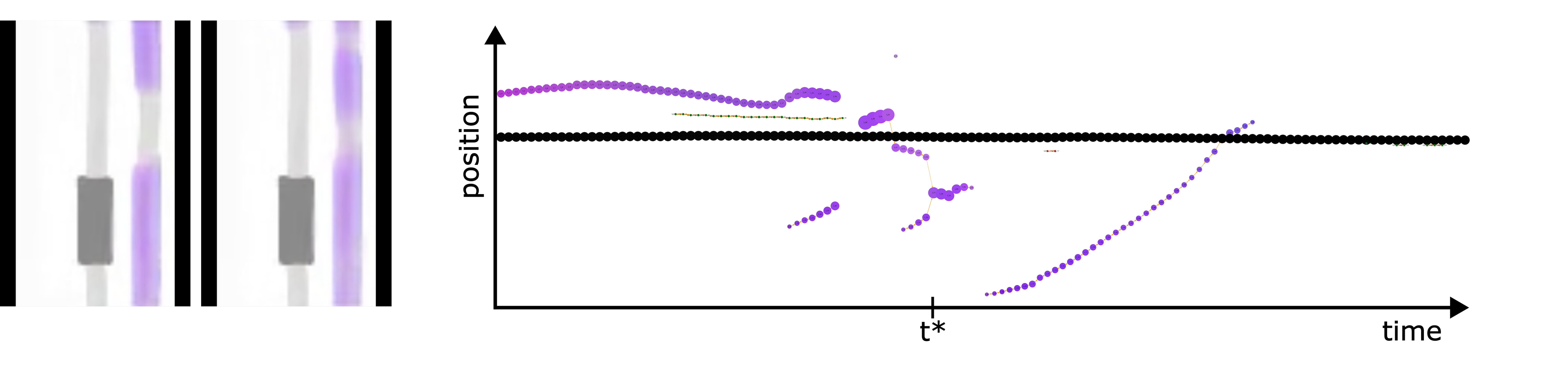}
    \caption{Left: Relevant frames from an example of merging hallucination observed at an early training stage. The frames are extracted around timestep t*. Right: the extracted graph of the objects in the scene. Notice that around t* two trajectories merge into one, causing one node to have two input nodes.}
    \label{fig:graph_trajectory_generated_hallucination}
\end{figure}

Analyzing the trajectory reconstruction graphs allows us to extract useful information about the quality of the generated videos. 
In particular, counting the amount of nodes in a scene which have an abnormal degree of input or output edges. 
A node with no parent node corresponds to an agent appearing in the scene, and a node with no children nodes corresponds to an agent disappearing. 
Similarly, a node with two parent nodes would correspond to a visual hallucination in which two distinct objects merge into one, and a node with two children nodes would correspond to an object splitting in two separate objects. These hallucinations are observed at early stages in training the videoGAN, and an example is provided in Figure~\ref{fig:graph_trajectory_generated_hallucination}. 

\begin{figure}[tb]
    \centering
    \includegraphics[width=0.8\textwidth]{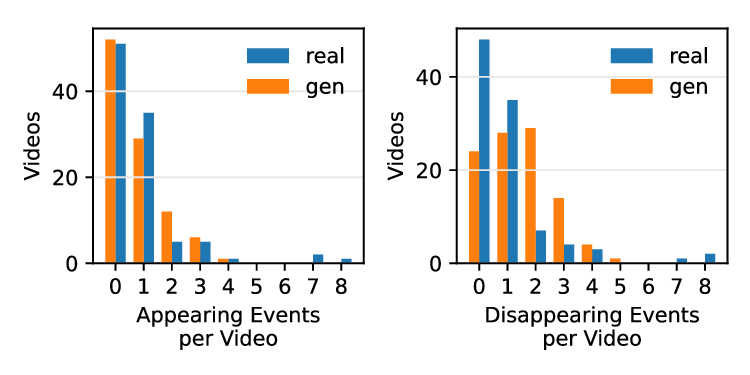}
    \caption{Distribution of appearing and disappearing events for FOV 15\,m, calculated over 100 real and 100 generated scenes.}
    \label{fig:hallucination_rates}
\end{figure}

As object permanence is one of the core difficulties of visual based prediction and generation, being able to track the occurrence of such hallucinations is very valuable in this type of work.
Computing the splitting and merging metrics on the well-performing checkpoints (at 20, 23 and 24 epochs respectively) over 100 scenes of 128 frames each yields rates lower than 1\% at FOV 15\,m and 20\,m, and marginally higher rates for FOV 25\,m (see Table~\ref{tab:hallucination_rates}).
Figure~\ref{fig:hallucination_rates} shows the distribution of scenes with different numbers of appearing and disappearing vehicle events for scenes of FOV 15\,m. The generated and real scene distributions are generally well aligned, with approximately half of the scenes containing no appearing events. A similar trend can be observed for disappearing events, although the generated scenes exhibit a slightly higher frequency of such events. 
It is important to note that the appearing and disappearing metrics do not directly measure the creation or removal of vehicles by the generative model. Instead, an appearing event can also occur when a vehicle enters the field of view of the ego vehicle, while a disappearing event can result from a vehicle leaving the observed area. Therefore, these events alone do not necessarily indicate hallucinations or violations of object permanence. The similarity between real and generated distributions suggests that the observed events primarily reflect realistic changes in scene visibility rather than artificial spawning or disappearance of agents.
\begin{table}[t]
\centering
\begin{tabular}{clcccc}
\toprule
& & \multicolumn{2}{c}{\textbf{Full frame}}
  & \multicolumn{2}{c}{\textbf{Boundary-excluded}} \\

\cmidrule(lr){3-4}
\cmidrule(lr){5-6}

\textbf{FOV} & \textbf{Category}
& \textbf{Generated} & \textbf{Real}
& \textbf{Generated} & \textbf{Real} \\

\midrule

15\,m
& Appearing    & 48.0\% & 49.0\% & 0.0\%  & 3.0\% \\
& Disappearing & 76.0\% & 52.0\% & 33.0\% & 2.0\% \\
& Merging      & 0.0\%  & 0.0\%  & 0.0\%  & 0.0\% \\
& Splitting    & 0.0\%  & 0.0\%  & 0.0\%  & 0.0\% \\

\addlinespace

20\,m
& Appearing    & 61.6\% & 58.6\% & 2.0\%  & 1.0\% \\
& Disappearing & 80.8\% & 56.6\% & 45.5\% & 1.0\% \\
& Merging      & 1.0\%  & 0.0\%  & 1.0\%  & 0.0\% \\
& Splitting    & 1.0\%  & 0.0\%  & 1.0\%  & 0.0\% \\

\addlinespace

25\,m
& Appearing    & 69.0\% & 70.7\% & 10.0\% & 1.0\% \\
& Disappearing & 87.0\% & 66.7\% & 55.0\% & 0.0\% \\
& Merging      & 4.0\%  & 1.0\%  & 2.0\%  & 0.0\% \\
& Splitting    & 6.0\%  & 1.0\%  & 2.0\%  & 0.0\% \\

\bottomrule
\end{tabular}

\vspace{4pt}

\caption{Percentage of videos containing at least one case of disappearing, appearing, merging or splitting events for different FOV, evaluation region ("Full frame" or "Boundary-excluded") and sample type (generated or real).}
\label{tab:hallucination_rates}
\end{table}
For this reason, we additionally report these metrics after excluding events occurring within the outer 10\% of the frame boundaries, where objects are likely to enter or leave the field of view. 
The right column ("Boundary-excluded") of Table~\ref{tab:hallucination_rates} counts the amount of videos that display at least one case of disappearing, appearing, merging or splitting vehicles that is not happening near the frame boundaries.
Overall, the generated scenes exhibit strong temporal consistency and realistic dynamics, with only occasional minor inconsistencies arising in highly static scenarios.

\subsection{Quantitative Results: Frame-level}

\begin{figure}[tb]
    \centering
    \subfloat[FOV 15\,m]{
        \includegraphics[width=0.32\linewidth]{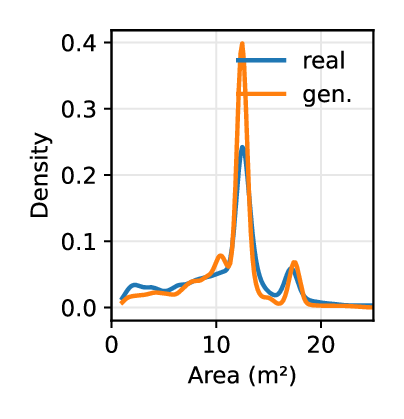}
    }
    \subfloat[FOV 20\,m]{
        \includegraphics[width=0.32\linewidth]{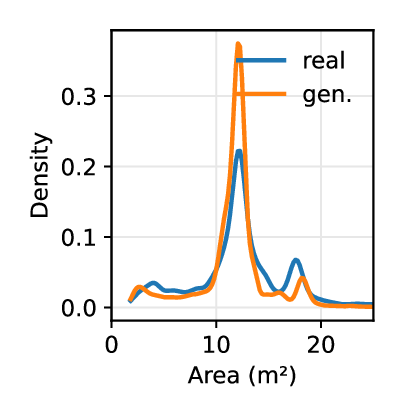}
    }
    \subfloat[FOV 25\,m]{
        \includegraphics[width=0.32\linewidth]{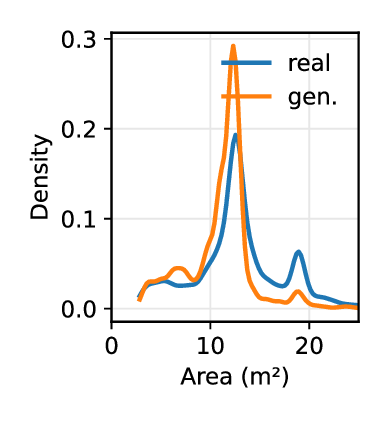}
    }
    \caption{Distribution of object footprint areas in real and generated scenes, for the three different investigated FOVs of 15\,m, 20\,m and 25\,m. The peak around 12\metersquare corresponds to the average area of a car.}
    \label{fig:object_areas}
\end{figure}

\begin{figure}[tb]
    \centering
    \subfloat[FOV 15\,m]{
        \includegraphics[width=0.3\linewidth]{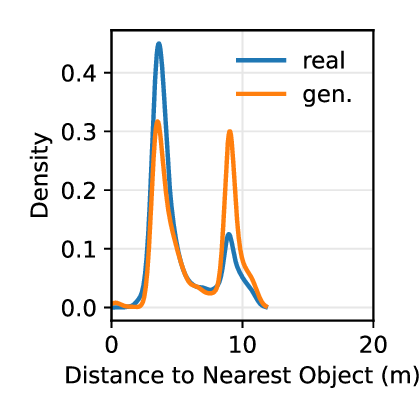}
    }
    \subfloat[FOV 20\,m]{
        \includegraphics[width=0.3\linewidth]{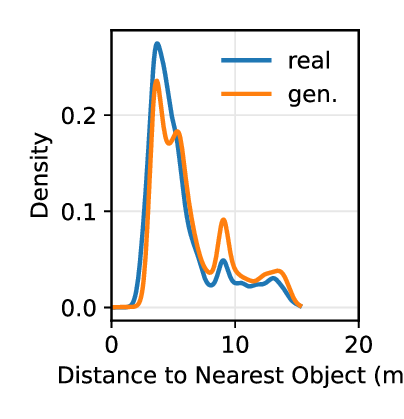}
    }
    \subfloat[FOV 25\,m]{
        \includegraphics[width=0.3\linewidth]{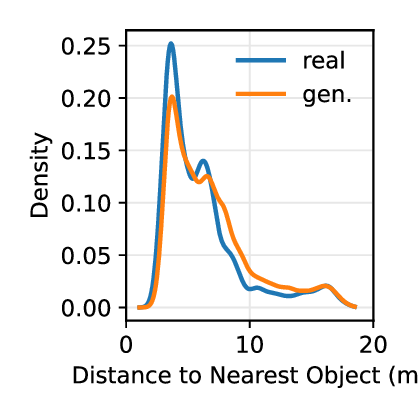}
    }
    \caption{Distribution of the distances between the two nearest vehicles in a scene, in real and generated scenes, for the three different investigated FOVs of 15\,m, 20\,m and 25\,m}
    \label{fig:nearest_object}
\end{figure}

We quantitatively evaluate the observations above by analyzing the distribution alignment for different geometrical parameters extracted from the generated and real videos.
We apply the same object detection pipeline to 200 real and 200 generated videos of duration 10\,s.
Figure~\ref{fig:object_areas} shows the distribution of the dimension of the extracted objects per each frame across the three investigated FOVs of 15\,m, 20\,m and 25\,m. The distribution over the generated videos matches the one over the real videos fairly well in all three cases.
The three modes of the distributions match the type of objects present in the scenes: some traffic lights and the occasional pedestrian around 2 to 5\metersquare, a great prevalence of cars around 12\metersquare, and some longer vehicle around 18\metersquare. Some of these longer vehicles are the result of inpainting a traffic light which is very near to a vehicle, as explained above. As observed above, this is mostly significant for vehicles waiting at a red traffic light.

\begin{figure}[tb]
    \centering
    \subfloat[FOV 15\,m]{
        \includegraphics[width=0.32\linewidth]{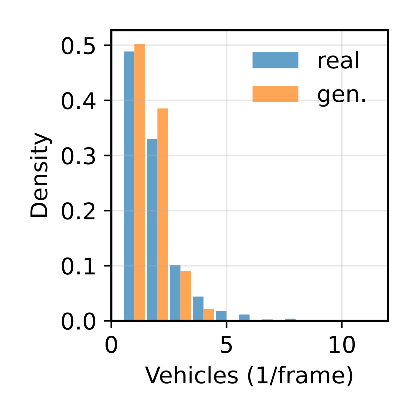}
    }
    \subfloat[FOV 20\,m]{
        \includegraphics[width=0.32\linewidth]{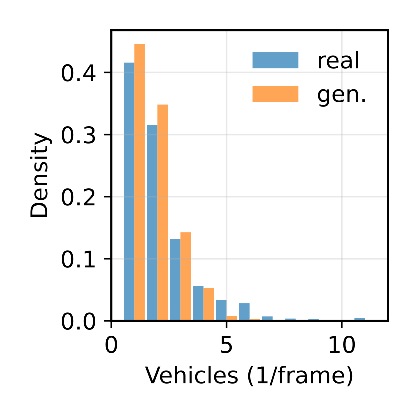}
    }
    \subfloat[FOV 25\,m]{
        \includegraphics[width=0.32\linewidth]{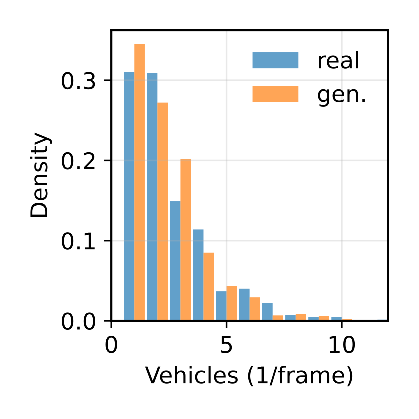}
    }
    \caption{Distribution of the amounts of vehicles in real and generated scenes, for the three different investigated FOVs of 15\,m, 20\,m and 25\,m. The amount of vehicles in scenes of FOV 15 ranges between 1 and 5, while it reaches up to 10 for scenes of FOV 25.}
    \label{fig:agents_per_frame}
\end{figure}

\begin{figure}[tb]
    \centering
    \subfloat[FOV 15\,m]{
        \includegraphics[width=0.32\linewidth]{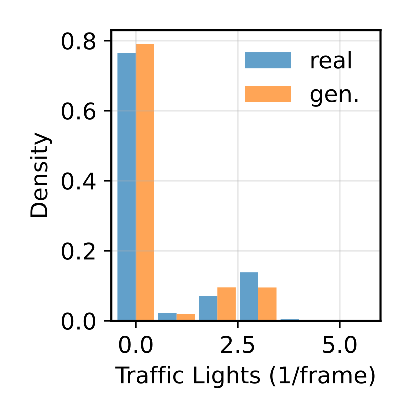}
    }
    \subfloat[FOV 20\,m]{
        \includegraphics[width=0.32\linewidth]{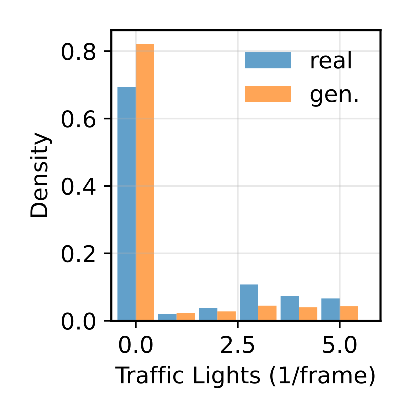}
    }
    \subfloat[FOV 25\,m]{
        \includegraphics[width=0.32\linewidth]{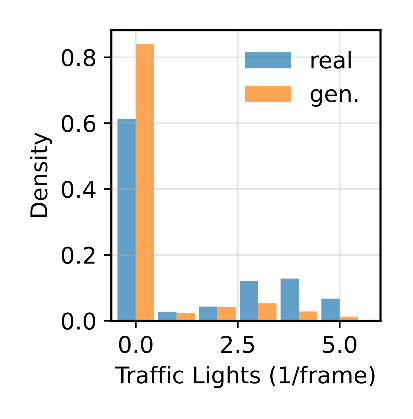}
    }
    \caption{Distribution of the amounts of traffic lights in real and generated scenes, for the three different investigated FOVs of 15\,m, 20\,m and 25\,m. In all three cases, most of the frames do not display traffic lights, and the ones who do have a preference for clustering in groups of around 3.}
    \label{fig:lights_per_frame}
\end{figure}

Figure~\ref{fig:nearest_object} shows the distribution of distances between the two nearest cars in a frame. Also in this case, the distributions match very well for all FOVs, with the generated scenes displaying even a marginally higher average distance between agents for FOV 15\,m. Overall, this suggests that generated road users maintain appropriate separation in the traffic. 
Finally, Figures~\ref{fig:agents_per_frame} and \ref{fig:lights_per_frame} report the average numbers of respectively vehicles and traffic lights observed in the generated and real videos. Also in this case the distributions are highly comparable, with the generated scenes presenting a marginally higher number of scenes with less traffic lights across all the FOVs. As expected, scenes with larger FOV encompass significantly more agents and traffic lights than scenes of lower FOV.

\subsection{Quantitative Results: Dynamics}

\begin{figure}[tb]
    \centering
    \subfloat[FOV 15\,m]{
        \includegraphics[width=0.32\linewidth]{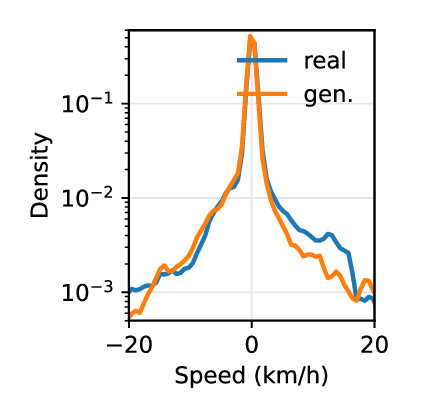}
    }
    \subfloat[FOV 20\,m]{
        \includegraphics[width=0.32\linewidth]{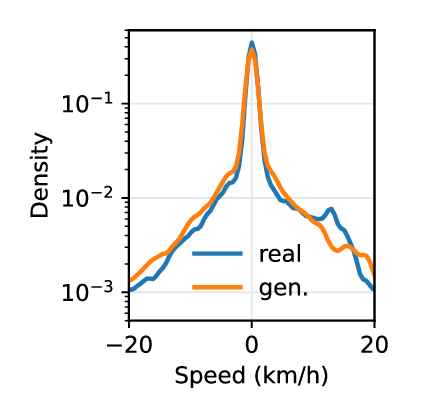}
    }
    \subfloat[FOV 25\,m]{
        \includegraphics[width=0.32\linewidth]{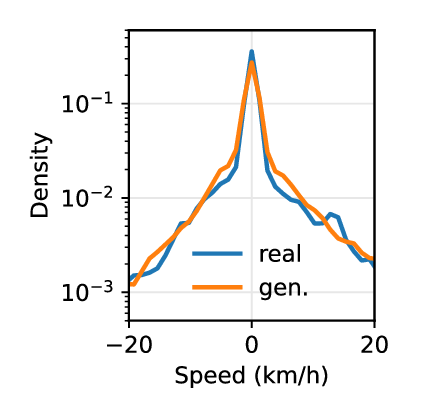}
    }
    \caption{Distribution of vehicle speeds in real and generated scenes, for the three different investigated FOVs of 15\,m, 20\,m and 25\,m, presented in logaritmic scale. The significant peak around 0 correspond to vehicles moving at the same speed as the ego vehicle (comoving traffic).}
    \label{fig:speeds}
\end{figure}

\begin{figure}[tb]
    \centering
    \subfloat[FOV 15\,m]{
        \includegraphics[width=0.32\linewidth]{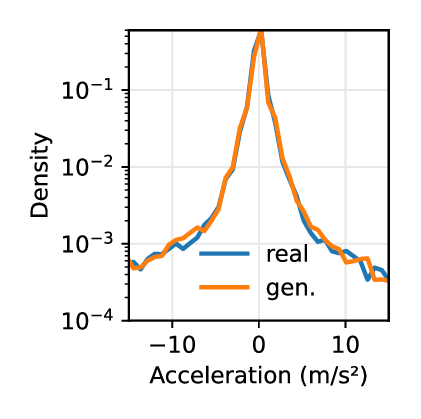}
    }
    \subfloat[FOV 20\,m]{
        \includegraphics[width=0.32\linewidth]{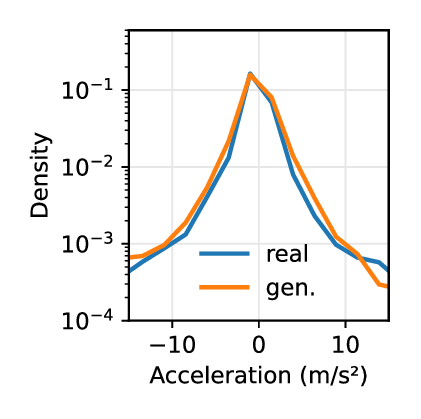}
    }
    \subfloat[FOV 25\,m]{
        \includegraphics[width=0.32\linewidth]{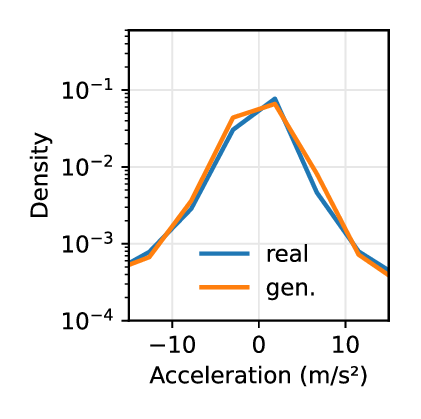}
    }
    \caption{Distribution of vehicle accelerations in real and generated scenes, for the three different investigated FOVs of 15\,m, 20\,m and 25\,m, presented in logaritmic scale.}
    \label{fig:accelerations}
\end{figure}

Figure~\ref{fig:speeds} and \ref{fig:accelerations} show the statistical distributions of relative speeds and accelerations of agents with the respect to the ego vehicle. Both distributions match very accurately also in logarithmic scale, which suggest that the video-generative model has learned the dynamics of the traffic scenes. This also confirms the qualitative observations reported above.
Positive speeds and accelerations are due to road users overtaking the ego vehicle, and similarly negative speeds are for agents moving in the opposite direction or being overtaken by the ego vehicle.
The sizable peak around 0 stems from co-moving traffic and the high prevalence of ego-only scenes.
This is also reflected in the broader distribution displayed by higher FOV scenes, which display a more diverse dynamic behavior in correlation with the higher amount of non-ego vehicles in each scene.

\begin{figure}[tb]
    \centering
    \subfloat[FOV 15\,m]{
        \includegraphics[width=0.32\linewidth]{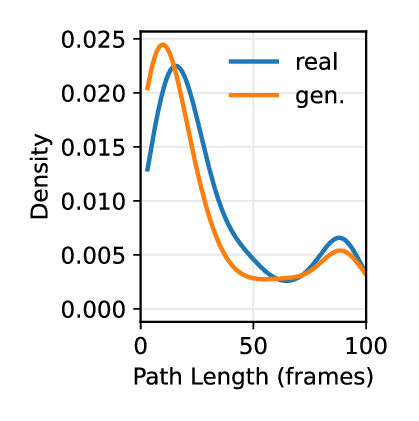}
    }
    \subfloat[FOV 20\,m]{
        \includegraphics[width=0.32\linewidth]{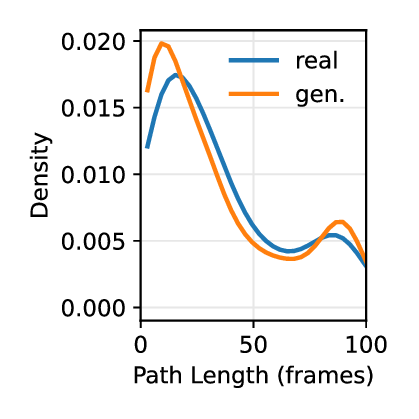}
    }
    \subfloat[FOV 25\,m]{
        \includegraphics[width=0.32\linewidth]{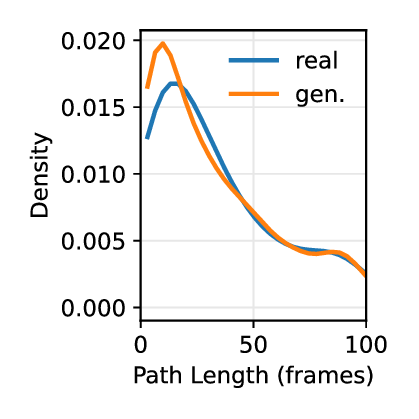}
    }
    \caption{Distribution of the lengths of real and generated trajectories as extracted from the videos for the three different investigated FOVs of 15\,m, 20\,m and 25\,m.}
    \label{fig:path_lengths}
\end{figure}

\begin{figure}[tb]
    \centering
    \subfloat[FOV 15\,m]{
        \includegraphics[width=0.32\linewidth]{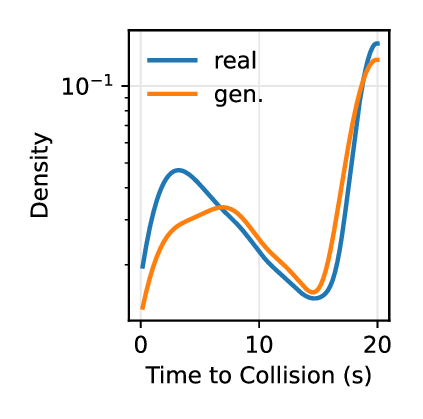}
    }
    \subfloat[FOV 20\,m]{
        \includegraphics[width=0.32\linewidth]{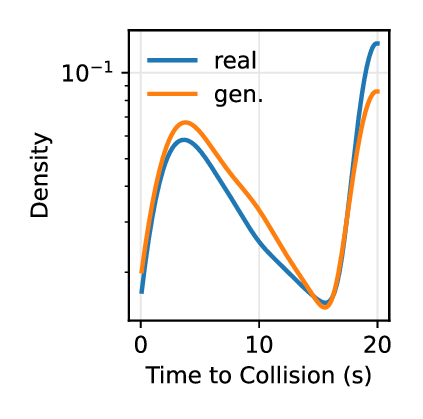}
    }
    \subfloat[FOV 25\,m]{
        \includegraphics[width=0.32\linewidth]{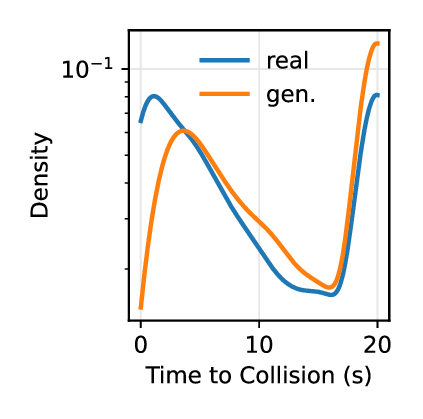}
    }
    \caption{Distribution of time-to-collision values for real and generated scenes, for the three different investigated FOVs of 15\,m, 20\,m and 25\,m.}
    \label{fig:time_to_collision}
\end{figure}

In Figure~\ref{fig:path_lengths} we observe that the average lengths of the trajectories extracted from the real and the generated videos are also comparable, although the real scenes display a slightly higher amount of medium-length scenes. This might correlate with the higher rate of agents leaving a scene observed in Figure~\ref{fig:hallucination_rates}.
Finally, we report the time-to-collision (TTC) metric,
computed as the ratio of the distance between two approaching agents and their relative distance, given in Figure~\ref{fig:time_to_collision}.
For visualization purposes only, TTC values exceeding 20s are displayed as 20s, as larger values indicate negligible collision risk and are less informative for comparison.
In all three cases, the generated scenes display a comparable or lower occurrence of low TTCs, which reflects on one side the absence of merging/colliding events reported above, and on the other it indicates once again that the generated trajectories do not look statistically less safe than the real one.

\subsection{Quantitative Results: Traffic Lights}
Since the model is unconditional, the generated map geometry may exhibit small local variations over time. To avoid introducing unnecessary uncertainty when estimating vehicle motion, we compute vehicle speeds relative to the traffic lights rather than with respect to the generated map itself. Qualitative inspection shows that traffic lights remain spatially consistent with the surrounding traffic throughout the generated sequences. We therefore treat them as static reference points and use the resulting relative vehicle speeds as a proxy for the absolute speeds when assessing realism.

\begin{figure}[tb]
    \centering
    \subfloat[FOV 15\,m]{
        \includegraphics[width=0.32\linewidth]{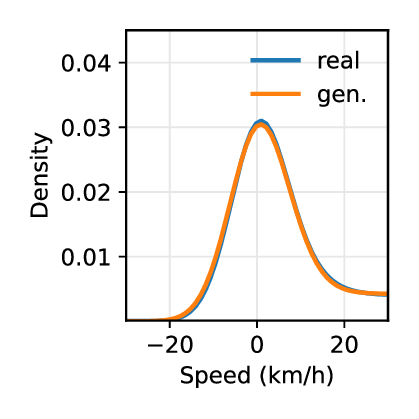}
    }
    \subfloat[FOV 20\,m]{
        \includegraphics[width=0.32\linewidth]{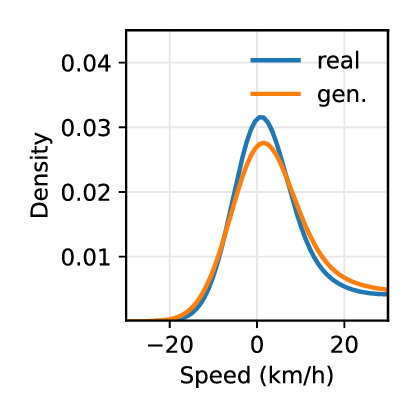}
    }
    \subfloat[FOV 25\,m]{
        \includegraphics[width=0.32\linewidth]{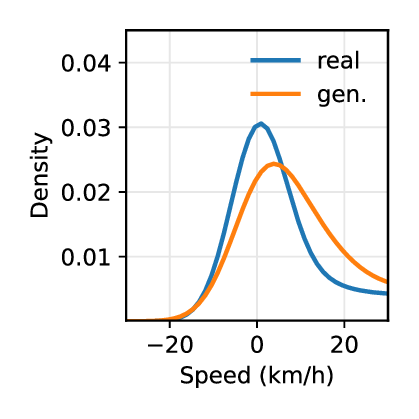}
    }
    \caption{Distribution of vehicle speeds in the proximity of a red light, in real and generated scenes, for the three different investigated FOVs of 15\,m, 20\,m and 25\,m.}
    \label{fig:red_lights}
\end{figure}

\begin{figure}[tb]
    \centering
    \subfloat[FOV 15\,m]{
        \includegraphics[width=0.32\linewidth]{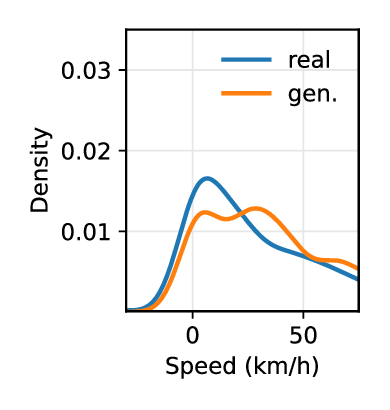}
    }
    \subfloat[FOV 20\,m]{
        \includegraphics[width=0.32\linewidth]{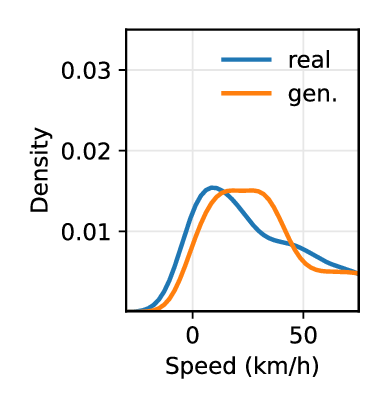}
    }
    \subfloat[FOV 25\,m]{
        \includegraphics[width=0.32\linewidth]{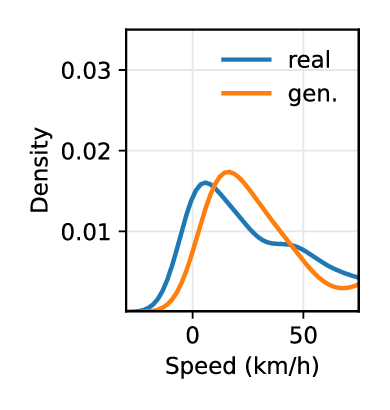}
    }
    \caption{Distribution of vehicle speeds in the proximity of a green light, in real and generated scenes, for the three different investigated FOVs of 15\,m, 20\,m and 25\,m.}
    \label{fig:green_lights}
\end{figure}

Figures~\ref{fig:red_lights} and ~\ref{fig:green_lights} report the distribution of the relative speed of the ego vehicle with respect to a traffic light in that frame.
For the red lights, in Figure~\ref{fig:red_lights}, we observe that in all cases the vehicle in the frame is most likely to be still.
This is particularly true for lower FOVs, while the higher FOVs seem to display a slightly higher probability for the agent to be moving.

Also for the distribution of speeds near green traffic lights, in Figure~\ref{fig:green_lights}, the generated distribution matches the real one in intensity and shape. The lower intensity of the peak at speed 0, which disappears for scenes of FOV 25, suggests that generated vehicles are slightly more likely to cross a green light, rather than staying still. However, they do not display higher rates of speeds over 50\,km/h when compared with the real scenes.
Overall the good match between the real and generated distribution suggests that the model learned in some measure a safe and realistic interaction with traffic lights, in agreement with visual observations.
Also observe that the range of speeds is compatible with average traffic speeds at large intersections like the ones present in the Waymo Dataset. This supports the validity of the hypothesis that traffic lights can be considered fixed in this analysis.

\section{Conclusions and Outlook}
In this work, we expanded upon \cite{mariani2025} to investigate a trajectory proposal pipeline based on a video generative model which is effective at learning the dynamics of traffic scenes in a purely visual fashion, without external rules or conditions. 
Trajectory data is encoded in 2D BEV semantic videos of traffic scenes, upon which a video generative model is trained, and finally trajectories are extracted from the generated videos.
In particular, we introduced improvements in the semantic encoding of the traffic scenes, which leads to better frame by frame object recognition in the trajectory extraction pipeline.
We also presented an improved graph-based object tracking method for semantic trajectory generation, and we used it to quantitatively assess the visual quality of generated scenes by measuring hallucinations, including disappearing and splitting objects.
We systematically investigated larger fields of view, evaluating scene generation over longitudinal ranges of 15\,m, 20\,m, and 25\,m, and we observed statistically safe and realistic trajectories over metrics such as speed, acceleration and time-to-collision distributions, with only slightly worsened statistics displayed in the larger FOV settings.
These results are observed while maintaining inference times below 20\,ms for generating traffic scenes of up to 20\,s, which are competitive for real-time downstream applications.

\begin{credits}

\subsubsection{\discintname}
The research leading to these results is funded by the German Federal Ministry for Economic Affairs and Energy within the project “NXT GEN AI METHODS – Generative Methoden für Perzeption, Prädiktion und Planung" (Grant no.\ 19A23014Q). The authors would like to thank the consortium for the successful cooperation.
The authors gratefully acknowledge the Gauss Centre for Supercomputing e.V. \url{https://www.gauss-centre.eu} for funding this project by providing computing time on the GCS Supercomputer JUWELS at Jülich Supercomputing Centre (JSC).
\end{credits}
%
%
%
\bibliographystyle{splncs04}
\bibliography{references}

\end{document}